%% file: main.tex
\pdfoutput=1

\documentclass[11pt]{article}

\usepackage[final]{acl}

\usepackage{times}
\usepackage{latexsym}

\usepackage[T1]{fontenc}

\usepackage[utf8]{inputenc}

\usepackage{microtype}

\usepackage{inconsolata}

\usepackage{graphicx}

\usepackage{amsmath}
\usepackage{booktabs}
\usepackage{multirow}
\usepackage{arydshln}
\usepackage{xspace}
\usepackage{algorithm}
\usepackage{algorithmic}
\newcommand{\proj}{MultiPruner\xspace}
\usepackage{hyperref}

\usepackage{xspace}
\usepackage{multirow}
\usepackage{amssymb}
\usepackage{multirow}
\usepackage{booktabs}

\usepackage{tabularray}
\usepackage{amsfonts}
\usepackage{pifont}
\usepackage{arydshln}
\usepackage{subcaption}
\newcommand{\xmark}{\ding{55}}%
\usepackage{mathtools}

\definecolor{customcolor1}{HTML}{E7DAD2}

\usepackage{tikz}
\usepackage{pifont}
\newcommand{\cmark}{\ding{51}}

\title{\proj: Balanced Structure Removal in Foundation Models}

\author{
 \textbf{J. Pablo Muñoz \textsuperscript{1}}\thanks{
 Co-first authors. 
 },
 \textbf{Jinjie Yuan\textsuperscript{2}}\footnotemark[1],
 \textbf{Nilesh Jain\textsuperscript{1}}
\\
 \textsuperscript{1}Intel Labs,
 \textsuperscript{2}Intel Corporation\\
 \small{
 \{pablo.munoz, jinjie.yuan, nilesh.jain\}@intel.com
 }
}

\begin{document}
\maketitle

\input{content/0_abstract}

\input{content/1_intro}

\input{content/2_methodology}

\input{content/3_experiments}

\input{content/4_related}

\input{content/5_conclusion}

\bibliography{main}

\clearpage

\appendix

\input{content/appendix}

\end{document}

%% file: content/0_abstract.tex
\begin{abstract}

Recently, state-of-the-art approaches for pruning large pre-trained models (LPMs) have demonstrated that the training-free removal of non-critical residual blocks in Transformers is viable for reducing model size, achieving results that outperform previous training-free pruning approaches.  
Motivated by these findings, we extend BlockPruner \cite{zhong2024blockprunerfinegrainedpruninglarge} and propose \textbf{\proj}, a pruning approach that surpasses recent training-free pruning methods by adopting a multidimensional, iterative, fine-grained pruning strategy.  
In \proj, multidimensional pruning reinstates the structural balance in block-pruned models by sequentially compressing along three dimensions: i) residual blocks, ii) channels of multilayer perceptrons (MLP), and iii) attention heads. This solution enhances zero-shot accuracy on downstream tasks compared to other techniques while improving model compression ratios, producing compressed models with fewer computing and memory requirements. Extensive experiments demonstrate the advantages of the proposed method across various large pre-trained models.
The code and pruning configurations are available at \href{https://github.com/IntelLabs/Hardware-Aware-Automated-Machine-Learning}{https://github.com/IntelLabs/Hardware-Aware-Automated-Machine-Learning}.

\end{abstract}

%% file: content/1_intro.tex
\begin{figure*}[ht]
    \centering    \includegraphics[width=\textwidth]{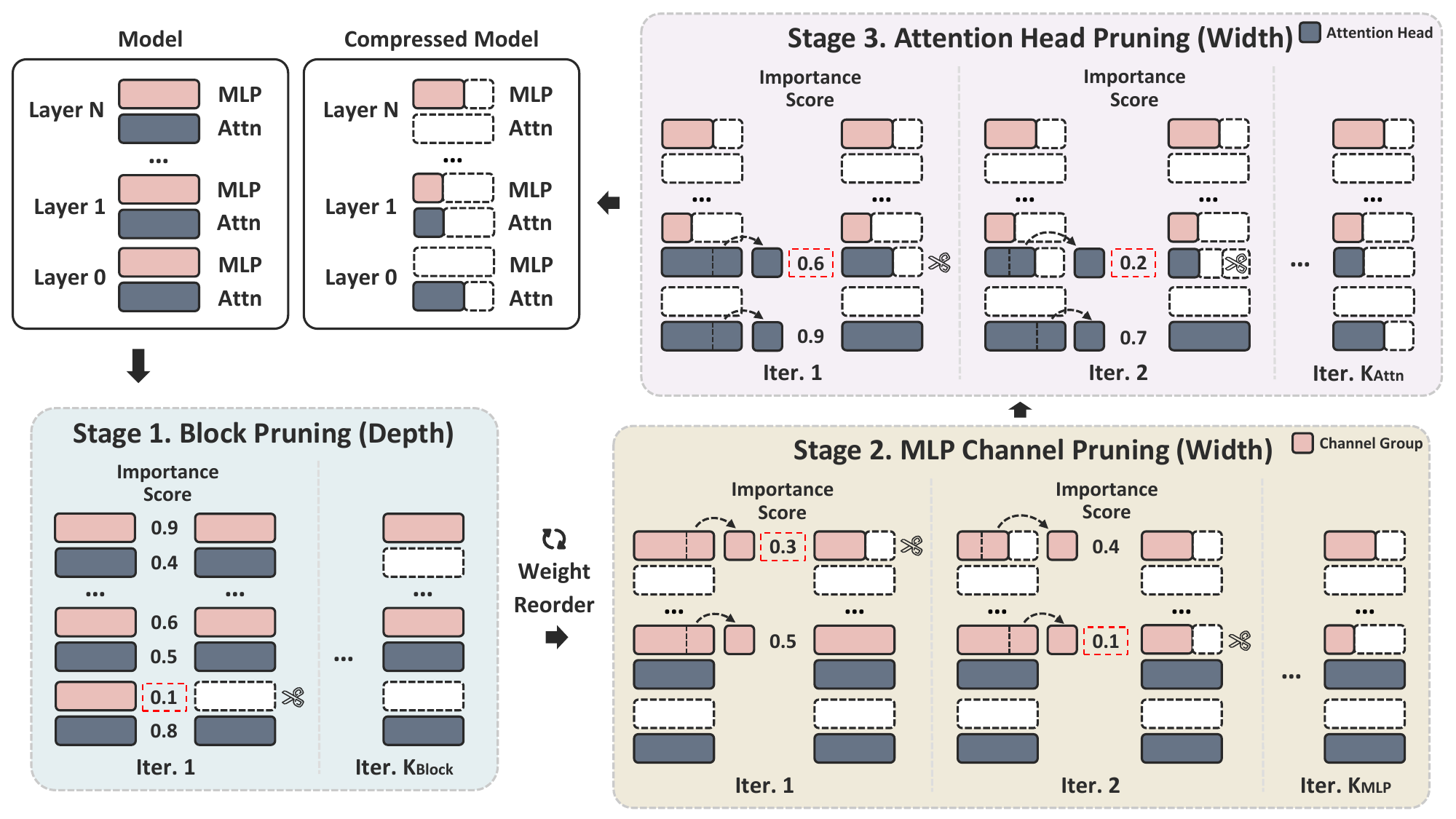}
    \caption{\proj adopts a multidimensional fine-grained pruning method to make pruning more balanced, resulting in a higher-performance pruned model.}
    \label{fig:algo}
\end{figure*}
\section{Introduction}

Large pre-trained models (LPMs), also known as foundation models (FMs) \cite{Bommasani2021FoundationModels}, such as GPT-4 \cite{openai2024gpt4technicalreport}, are producing outstanding results across a variety of domains, and have motivated an increase in investment in artificial intelligence (AI)-related ventures. State-of-the-art AI models often have billions of parameters and require large clusters of graphics processing units (GPUs) to train. Pre-trained models are usually subject to a less resource-intensive subsequent stage in which the model is adapted or fine-tuned for a downstream task. Beyond the challenges of training and fine-tuning, deploying these models requires complex systems with significant computing and memory capacity. Two delimited stages are in play during inference: prefill and decode. In the prefill stage, all the required caches are created, which tends to be compute-bound. In the decode stage, the model uses the existing caches to generate new tokens, which tends to be memory-bound.  

Given the substantial resource requirements for training, fine-tuning, and deploying these models, model compression techniques have become increasingly important. 
Pruning and quantizing LPMs have been proposed to reduce resource requirements, improve model performance, and enable deployment to more limited environments. However, choosing a particular compression algorithm requires considering the large parameter space of these models and the cost in time and resources that this process will take. For instance, when utilizing pruning, the algorithm should be capable of efficiently analyzing which model components might be removed, guaranteeing a sustained performance, e.g., with a minor tolerable drop in accuracy.  

Recently, ShortGPT \cite{men2024shortgptlayerslargelanguage} demonstrated that removing Transformer blocks based on their relative importance is a viable approach for pruning large pre-trained models. BlockPruner \cite{zhong2024blockprunerfinegrainedpruninglarge} took this idea further to demonstrate that Transformer blocks can be partitioned into their two sub-components based on the residual connections, i.e., minimal residual blocks and showed that this more fine-grained pruning approach results in models with higher a pruning ratio and accuracy. 
However, these works assume that structured pruning should be applied to the Transformer's depth dimension, which can easily lead to over-pruning in a single dimension, removing necessary layers or blocks. This paper considers these challenges and extends BlockPruner \cite{zhong2024blockprunerfinegrainedpruninglarge}, resulting in a method to advance the state-of-the-art in structured pruning of large pre-trained models. \proj removes this assumption and demonstrate that block pruning can be complemented with additional pruning in other dimensions, maintaining the target efficiency. Our approach, \proj, is a training-free approach that extends the benefits of state-of-the-art training-free block pruning approaches and produces smaller models with higher accuracy. Specifically, \proj operates in three pruning stages. First, it prunes the least important residual blocks, leveraging the insights from BlockPruner \cite{zhong2024blockprunerfinegrainedpruninglarge} to identify minimal residual blocks that can be removed without significantly impacting performance. Next, it applies a fine-grained pruning strategy to the MLP channels, followed by the attention heads, aiming to prune the model more accurately.  
This sequential pruning process ensures that each dimension is optimally compressed, resulting in a smaller and more efficient model.

In summary, \proj advances training-free model compression and includes the following contributions:
\begin{enumerate}
    \item A pruning algorithm, \proj, that extends BlockPruner \cite{zhong2024blockprunerfinegrainedpruninglarge} and provides an iterative fine-grained pruning strategy across multiple dimensions, leading to enhanced zero-shot accuracy on downstream tasks and improved model compression rates. 
    \item Studies to explore the challenges in multidimensional pruning, such as the pruning order and ratios for different types of structures.
    \item Extensive experiments demonstrating the effectiveness of \proj across various pruning ratios and models. 
\end{enumerate}

The following sections provide a comprehensive overview of the benefits and applications of \proj, organized as follows: Section \ref{sec:method} discusses the proposed method followed by Section \ref{sec:experiments} with experimental results. Related work is discussed in Section \ref{sec:related_w}. We conclude with our thoughts on the impact of our research.

%% file: content/2_methodology.tex
\section{Methodology}
\label{sec:method}
 
\proj is motivated by recent block pruning algorithms, e.g., \citet{zhong2024blockprunerfinegrainedpruninglarge, men2024shortgptlayerslargelanguage}, that remove elements on a single dimension. 
Based on their importance, these methods prune Transformer blocks or their sub-components, like self-attention or multilayer perceptron (MLP) blocks. However, focusing solely on these coarse residual blocks might leave pruning opportunities unrealized in other dimensions. 
\proj adopts a fine-grained approach to structural pruning while exploring the structural balance of the given architecture, since by removing complete residual blocks in Transformers, we are altering the original network design and the choices that might have been determined after expensive exploration and experimentation by the model's creators. For instance, a pruned network becomes shallower after applying block pruning, but its overall width remains unchanged. By pruning in other orthogonal dimensions, \proj attempts to reinstate this balance, resulting in compressed high-performing models that closely follow the original design considerations. The benefits of \proj are demonstrated by experiments results in Section \ref{sec:experiments}.  

\begin{algorithm}[H]
  \caption{Multidimensional Pruning with Fixed Thresholds per Pruning Target}
  \label{alg:fixed_pruning}   
  \begin{algorithmic}[1]
    \small
    \REQUIRE Set of minimal residual blocks $\mathcal{M}$ from a model $m$, s.t. $\mathcal{M} = \{M_i \mid M_i \in \mathcal{M}, \text{type}(M_i) \in \{\text{MLP}, \text{ATTN}\}\}$, Calibration dataset $\mathcal{C}$, Metric $\phi$, Target pruning ratios $\tau_1$, $\tau_2$, and $\tau_3$ for each pruning modality, MLP channel group size $g_{\text{MLP}}$, ATTN channel group size $g_{\text{ATTN}}$.
    \ENSURE Pruned model $m^*$
    \STATE $\tau \gets 0$ \\
    
    \WHILE{$\tau < \tau_1$}
        \FORALL{$M_i \in \mathcal{M}$}
            \STATE $S_{i} \gets \text{BlockImportance}(M_i, m, \mathcal{C}, \phi)$
        \ENDFOR
        \STATE $M_{\text{min}} \gets \arg\min_{M_i \in \mathcal{M}} S_{i}$
        \STATE $\mathcal{M} \gets \mathcal{M} \setminus \{M_{min}\}$
        \STATE $\tau \gets \text{PruningRatio}(m)$
    \ENDWHILE
    \STATE $\text{WeightReordering}(m)$
    \FORALL{$t \in \{\text{MLP}, \text{ATTN}\}, \tau_t \in \{\tau_2, \tau_3\}, g_t \in \{g_{\text{MLP}}, g_{\text{ATTN}}\}$}
        \STATE $\mathcal{M}_{t} = \{M_i \mid M_i \in \mathcal{M}, \text{type}(M_i) == t \}$ 
        \WHILE{$\tau < \tau_t$}
            \FORALL{$M_i \in \mathcal{M}_{t}$}
                \STATE $S_{i} \gets \text{WidthImportance}(M_i[:, \text{:-}g_{t}], m, \mathcal{C}, \phi)$
            \ENDFOR 
            \STATE $M_{\text{min}} = \arg\min_{M_i \in \mathcal{M}_{t}} S_i$
            \STATE $M_{\text{min}} = M_{\text{min}}[:, \text{:-}g_{t}]$ 
            \STATE $\tau \gets \text{PruningRatio}(m)$
        \ENDWHILE
    \ENDFOR
    
    \RETURN $m^* \text{ with the remaining and altered blocks in } \mathcal{M}$
  \end{algorithmic}
\end{algorithm}

\paragraph{Problem Definition}
Given a dense model $m$, e.g., Llama-2-7B \cite{touvron2023llama}, associated with a set of minimum residual blocks $\mathcal{M}$ from each Transformer block (self-attention or MLP), and a target pruning ratio $\tau$, \proj takes a finer-grained pruning approach compared to other pruning solutions to obtain a model $m^*$ with an associated subset of altered blocks from $\mathcal{M}$. \textbf{\proj's objective is to find a high-performing pruning configuration that results in a similar pruning ratio to the competing state-of-the-art block pruning while maintaining a structural balance and improving its zero-shot performance on downstream tasks.}   

\subsection{Multidimensional Fine-Grained Pruning} 
\proj targets three main elements for structure removal: 
\begin{itemize}
    \item Residual Transformer Blocks (Depth)
    \item MLP Channels (Width)
    \item Attention Heads (Width)
\end{itemize}

In the depth dimension, \proj removes \emph{iteratively} the least important minimal residual blocks as in BlockPruner \cite{zhong2024blockprunerfinegrainedpruninglarge}. 
In the width dimension, \proj removes groups of channels from the MLP and attention heads. When exploring various architectural dimensions of the model for pruning and considering that a full search of architecture configurations is not practical for large pre-trained models, even when using training-free approaches and zero-shot evaluation, a natural research question arises: 
\\~\\
\noindent
\emph{(1) Should \proj prune the model's depth and width in parallel or sequentially to obtain a high-performing pruned model?}
\\~\\
The parallel strategy adds complexity and does not provide insights regarding the contributions of pruning in each dimension. As detailed in Section \ref{sec:experiments}, experimentally, we have observed that following sequential steps to prune different model's dimensions yields the best pruned models. However, if the decision is to prune each dimension sequentially, we are confronted with a second research question:
\\~\\
\noindent
\emph{(2) What is the recommended order to sequentially explore the removal of structures in each dimension?}
\\~\\
Intuitively, a coarse-to-fine-grained order, i.e., from blocks to MLP channels to attention heads, will result in a more precise pruned model. This intuition is supported by our experimental results in Section \ref{sec:experiments}, which demonstrate that \proj achieves better performance when it first removes structures along the depth of the model. Once this stage is completed, it focuses on structures along the width of selected components. As shown in Figure \ref{fig:algo}, \proj begins by pruning the least important residual Transformer blocks, reducing the model's depth. Following this, it targets the MLP Channels, and finally, it prunes the attention heads, which raises an additional research question: 
\\~\\
\noindent
\emph{(3) When must \proj stop pruning in each dimension or type of component?}
\\~\\
To answer this last question, we assign each of the three pruning targets with a pruning ratio threshold, $\tau_1, \tau_2, $ and $\tau_3$ (i.e., the target ratio $\tau$). To discover values for these thresholds that yield high-performing models, we utilize two search strategies: 

\begin{itemize}
    \item Fixed thresholds per pruning target. 
    \item Fixed threshold for the depth dimension and evolutionary search to discover Pareto-optimal configurations on the width dimension.
\end{itemize}

Next, we discuss these search variants in more detail. 

\begin{table*}[!t]
    \setlength{\tabcolsep}{3pt}
    \centering
    \footnotesize
    \begin{tabular}{llcccccccc}
    \toprule
        \textbf{Model} & \textbf{Method} & \textbf{Ratio (\%)} & \textbf{PPL ($\downarrow$)} &\textbf{PIQA} & \textbf{WinoG} & \textbf{HellaS} & \textbf{ARC-e} & \textbf{ARC-c} & \textbf{Avg. Score} \\ 
    \midrule
        \multirow{7}{*}{\textbf{Llama2-7B}} & Dense & 0 & 5.47 & 79.05  & 69.06  & 75.99  & 74.54  & 46.16  & 68.96   \\ 
          ~ & SliceGPT & 21.45 & 30.74 & 72.42  & 59.91  & 56.04  & 63.64  & 37.12  & 57.83   \\ 
          ~ & LaCo & 21.02 & 50.39 & 68.34  & 60.46  & 54.08  & 55.39  & 35.84  & 54.82   \\ 
          ~ & RM & 21.02 & 676.80 & 54.46  & 49.25  & 29.22  & 34.43  & 22.53  & 37.98   \\ 
          ~ & ShortGPT & 21.02 & 18.45 & 70.24  & \textbf{65.90}  & 62.63  & 56.06  & 36.09  & 58.18   \\ 
          ~ & BlockPruner & 21.99 &  11.51 & 74.21 & 62.43  & 65.87  & 61.07  & 37.29  & 60.17   \\ 
          ~ & \cellcolor{customcolor1}\textbf{\proj} & \cellcolor{customcolor1}21.96 &  \cellcolor{customcolor1}\textbf{9.33} & \cellcolor{customcolor1}\textbf{74.65} & \cellcolor{customcolor1}64.64 & \cellcolor{customcolor1}\textbf{68.94} & \cellcolor{customcolor1}\textbf{64.77} & \cellcolor{customcolor1}\textbf{41.13} & \cellcolor{customcolor1}\textbf{62.83} \\
    \midrule 
         \multirow{7}{*}{\textbf{Llama2-13B}} & Dense & 0 & 4.89 & 80.52  & 72.14  & 79.36  & 77.36  & 49.23  & 71.72   \\ 
        ~ & SliceGPT & 21.52 & 23.95 & 74.32  & 65.59  & 60.71  & 68.52  & 42.41  & 62.31   \\ 
        ~ & LaCo & 24.37 & 13.97 & 72.42  & 59.27  & 60.44  & 54.34  & 34.56  & 56.21   \\ 
        ~ & RM & 24.37 & 10.08 & 73.72  & 66.61  & 66.80  & 66.12  & 41.98  & 63.05   \\ 
        ~ & ShortGPT & 24.37 & 20.06 & 72.74  & 70.80  & 67.80  & 60.35  & 41.30  & 62.60   \\ 
        ~ & BlockPruner & 25.12 &  8.16 & 76.93  & 66.30  & 72.20  & 65.82  & 41.38  & 64.53   \\ 
        ~ & \cellcolor{customcolor1}\textbf{\proj} & \cellcolor{customcolor1}25.01 &  \cellcolor{customcolor1}\textbf{7.19} & \cellcolor{customcolor1}\textbf{77.80} & \cellcolor{customcolor1}\textbf{71.90} & \cellcolor{customcolor1}\textbf{75.62} & \cellcolor{customcolor1}\textbf{71.38} & \cellcolor{customcolor1}\textbf{46.50} & \cellcolor{customcolor1}\textbf{68.64} \\ 
    \midrule
        \multirow{6}{*}{\textbf{Baichuan2-7B}} & Dense & 0 & 6.04 & 77.48  & 68.27  & 72.18  & 72.98  & 42.75  & 66.73   \\ 
        ~ & LaCo & 21.57 & 26.46 & 68.28  & 58.56  & 51.50  & 52.90  & 28.50  & 51.95   \\ 
        ~ & RM & 21.57 & 189.78 & 59.96  & 52.33  & 30.87  & 38.17  & 23.63  & 40.99   \\ 
        ~ & ShortGPT & 21.57 & 31.05 & 63.71  & 62.67  & 50.01  & 47.31  & 30.72  & 50.88   \\ 
        ~ & BlockPruner & 22.45 &  15.38 & 69.75  & 61.48  & 58.09  & \textbf{58.08}  & 33.02  & 56.08   \\ 
        ~ & \cellcolor{customcolor1}\textbf{\proj} & \cellcolor{customcolor1}22.41 &  \cellcolor{customcolor1}\textbf{12.37} & \cellcolor{customcolor1}\textbf{70.46} & \cellcolor{customcolor1}\textbf{64.72} & \cellcolor{customcolor1}\textbf{60.92} & \cellcolor{customcolor1}56.99 & \cellcolor{customcolor1}\textbf{34.90} & \cellcolor{customcolor1}\textbf{57.60}  \\
    \midrule
    \multirow{6}{*}{\textbf{Baichuan2-13B}} & Dense & 0 & 6.66 & 78.84  & 70.40  & 75.23  & 74.07  & 47.70  & 69.25   \\ 
        ~ & LaCo & 22.68 & 27.07 & 70.89  & 58.01  & 54.00  & 57.11  & 32.94  & 54.59   \\ 
        ~ & RM & 22.68 & 17.70 & 68.99  & 67.88  & 63.78  & 57.49  & 37.54  & 59.14   \\ 
        ~ & ShortGPT & 22.68 & 20.69 & 69.31  & \textbf{68.27}  & 61.71  & 56.52  & 36.69  & 58.50   \\ 
        ~ & BlockPruner & \cellcolor{white}24.19 &  15.36 & \textbf{71.44}  & 64.01  & \textbf{64.20}  & \textbf{59.81}  & \textbf{37.88}  & \textbf{59.47}   \\
        ~ & \cellcolor{customcolor1}\textbf{\proj} &\cellcolor{customcolor1}24.01	&\cellcolor{customcolor1}\textbf{10.99}	&\cellcolor{customcolor1}69.97	&\cellcolor{customcolor1}66.46&	\cellcolor{customcolor1}64.03	&\cellcolor{customcolor1}57.95	&\cellcolor{customcolor1}37.20	&\cellcolor{customcolor1}59.12 \\
    \midrule
        \multirow{6}{*}{\textbf{Qwen1.5-7B}} & Dense & 0 & 7.95 & 79.22  & 66.46  & 76.92  & 62.16  & 42.66  & 65.48   \\ 
        ~ & LaCo & 20.97 & 39.23 & 70.40  & 58.64  & 56.35  & 46.89  & 32.85  & 53.03   \\ 
        ~ & RM & 20.97 & 2026.31 & 67.36  & 49.88  & 42.00  & 54.17  & 28.58  & 48.40   \\ 
        ~ & ShortGPT & 20.97 & 49.88 & 69.53  & \textbf{62.12}  & 58.87  & 43.60  & 32.17  & 53.26   \\ 
        ~ & BlockPruner & 21.83 &  20.58 & 71.71  & 55.56  & 59.31  & 53.70  & 33.28  & 54.71   \\ 
        ~ & \cellcolor{customcolor1}\textbf{\proj} & \cellcolor{customcolor1}21.81 & \cellcolor{customcolor1}\textbf{18.22} & \cellcolor{customcolor1}\textbf{71.76} & \cellcolor{customcolor1}59.59 & \cellcolor{customcolor1}\textbf{60.60} & \cellcolor{customcolor1}\textbf{59.01} & \cellcolor{customcolor1}\textbf{35.92} & \cellcolor{customcolor1}\textbf{57.38} \\
    \midrule
        \multirow{6}{*}{\textbf{Qwen1.5-14B}} & Dense & 0 & 7.44 & 79.87  & 70.56  & 79.41  & 68.48  & 47.01  & 69.07   \\ 
        ~ & LaCo & 22.25 & 16.32 & 71.55  & 58.33  & 60.16  & 53.70  & 34.04  & 55.56   \\ 
        ~ & RM & 22.25 & 55.99 & 67.08  & 53.28  & 42.08  & 50.72  & 29.01  & 48.43   \\ 
        ~ & ShortGPT & 22.25 & 1237.21 & 58.60  & 55.96  & 36.16  & 38.09  & 34.81  & 44.72   \\ 
        ~ & BlockPruner & 23.72 &  15.67 & 75.24  & 61.48  & \textbf{66.92}  & 59.51  & 39.08  & 60.45   \\
        ~ & \cellcolor{customcolor1}\textbf{\proj} & \cellcolor{customcolor1}23.51 &  \cellcolor{customcolor1}\textbf{12.94} & \cellcolor{customcolor1}\textbf{75.41} & \cellcolor{customcolor1}\textbf{62.98} & \cellcolor{customcolor1}63.26 & \cellcolor{customcolor1}\textbf{69.19} & \cellcolor{customcolor1}\textbf{41.21} & \cellcolor{customcolor1}\textbf{62.41} \\ 
    \bottomrule
    \end{tabular}
\caption{
Zero-shot downstream task performance of various models using different pruning methods. ``Dense'' denotes the original (unpruned) models. ``PPL'' refers to the perplexity in Wikitext2. PIQA, WinoG, HellaS, ARC-e, and ARC-c represent their respective accuracies.
}   
\label{tab:main_res}
\end{table*}

\begin{table}[!t]
    \setlength{\tabcolsep}{1.5pt}
    \centering
    \footnotesize
    \renewcommand*\arraystretch{1.1}
    \begin{tabular}{llccc}
    \toprule
        \textbf{Model} & \textbf{Method} & \textbf{Ratio (\%)} & \textbf{PPL ($\downarrow$)} & \textbf{Avg. Score} \\ 
    \midrule
        \multirow{3}{*}{\textbf{Llama3.2-3B}} & Dense & 0 & 7.81 & 67.67   \\ 
          ~ & BlockPruner & 9.39 &  13.07  & 62.31   \\ 
          ~ & \cellcolor{customcolor1}\textbf{\proj} & \cellcolor{customcolor1}9.06 &  \cellcolor{customcolor1}\textbf{10.46} & \cellcolor{customcolor1}\textbf{64.04} \\
    \midrule
        \multirow{5}{*}{\textbf{Llama3.1-8B}} & Dense & 0 & 6.24 & 73.75   \\ 
          ~ & BlockPruner & 10.65 &  10.58  & 66.75   \\ 
          ~ & \cellcolor{customcolor1}\textbf{\proj} & \cellcolor{customcolor1}10.03 &  \cellcolor{customcolor1}\textbf{8.93} & \cellcolor{customcolor1}\textbf{69.27} \\
          \cdashline{2-5}
          ~ & BlockPruner & 19.95 &  15.37  & 59.08   \\ 
          ~ & \cellcolor{customcolor1}\textbf{\proj} & \cellcolor{customcolor1}20.00 &  \cellcolor{customcolor1}\textbf{13.86} & \cellcolor{customcolor1}\textbf{63.07} \\
    \midrule
        \multirow{5}{*}{\textbf{Llama3-8B}} & Dense & 0 & 6.14 & 72.73   \\ 
          ~ & BlockPruner & 10.13 &  10.88  & 66.46   \\ 
          ~ & \cellcolor{customcolor1}\textbf{\proj} & \cellcolor{customcolor1}10.08 &  \cellcolor{customcolor1}\textbf{8.19} & \cellcolor{customcolor1}\textbf{69.03} \\
          \cdashline{2-5}
          ~ & BlockPruner & 20.47 &  22.36  & 57.59   \\ 
          ~ & \cellcolor{customcolor1}\textbf{\proj} & \cellcolor{customcolor1}20.00 &  \cellcolor{customcolor1}\textbf{16.01} & \cellcolor{customcolor1}\textbf{63.02} \\
    \midrule
        \multirow{5}{*}{\textbf{Qwen2.5-7B}} & Dense & 0 & 6.85 & 72.04   \\ 
          ~ & BlockPruner & 10.34 &  9.88  & 67.44   \\ 
          ~ & \cellcolor{customcolor1}\textbf{\proj} & \cellcolor{customcolor1}10.02 &  \cellcolor{customcolor1}\textbf{9.15} & \cellcolor{customcolor1}\textbf{69.71} \\
          \cdashline{2-5}
          ~ & BlockPruner & 20.29 &  17.17  & 57.44   \\ 
          ~ & \cellcolor{customcolor1}\textbf{\proj} & \cellcolor{customcolor1}20.02 &  \cellcolor{customcolor1}\textbf{13.37} & \cellcolor{customcolor1}\textbf{62.82} \\
    \bottomrule
    \end{tabular}
\caption{
Zero-shot downstream task performance of more advanced LLMs compared to BlockPruner. ``Dense'' denotes the original (unpruned) models. ``PPL'' refers to the perplexity in Wikitext2. Average score means the average accuracy across five downstream tasks. 
}   
\label{tab:main_res2}
\end{table}

\paragraph{Sequential Pruning with Fixed Targets}

Experimentally, we discover the value for hyper-parameters, $\tau_1, \tau_2, $ and $\tau_3$ that determine the pruning ratio of each type of component: complete residual blocks, MLP channels, and attention heads, respectively.  
The entire process is detailed in Algorithm \ref{alg:fixed_pruning} and Figure \ref{fig:algo}.  

The algorithm begins by pruning the depth dimension, where \proj removes the least important residual Transformer blocks iteratively until the pruning ratio $\tau_1$ is reached (lines 2-9 in Algorithm \ref{alg:fixed_pruning}). 
The importance metric used for all pruning steps is the perplexity (PPL) on the calibration dataset.
This step ensures that the model's depth is reduced in a controlled manner, preserving the most critical blocks based on their importance scores.

After completing the depth pruning, \proj performs a weight reordering step (line 10 in Algorithm \ref{alg:fixed_pruning}). This step aims to prioritize the less important channels and heads for pruning. By reordering the weights based on an importance metric, the channels and heads are reordered such that the least important ones are positioned last, making them the primary candidates for pruning. In our main experiment, we employed the L1 Norm as the reorder metric, and more details can be found in Appendix \ref{sec:weight_reordering}.

Once the weight reordering is complete, \proj shifts focus to the width dimension. It sequentially prunes non-essential groups of channels in the MLP and attention heads, adhering to the pruning ratios $\tau_2, $ and $\tau_3$, respectively (lines 11-21 in Algorithm \ref{alg:fixed_pruning}). The algorithm calculates the importance scores for each component type and removes the least important channel group or head iteratively until the target pruning ratio is achieved.

This sequential approach ensures that each dimension is pruned effectively, balancing depth and width reduction. By following this method, \proj achieves a high-performing pruned model that closely aligns with the original design considerations while significantly reducing the model's size.

\paragraph{Fixed Threshold for Depth and Evolutionary Search for Width}
After removing residual blocks in the depth pruning stage, we have experimentally observed that stopping at half of the pruning ratio, i.e., $\tau/2$, provides an effective heuristic to initiate the pruning using evolutionary search in the width dimension. This stopping point opens a significant opportunity to remove elements in the width dimension, resulting in a better balance in the model.  

\begin{algorithm}
  \caption{Multidimensional Pruning with Fixed Delimitation and Evolutionary Search.
  }
  \label{alg:evol_pruning}
  \begin{algorithmic}[1]
  \small
  \REQUIRE Set of minimal residual blocks $\mathcal{M}$ from a model $m$, s.t. $\mathcal{M} = \{M_i \mid M_i \in \mathcal{M}, \text{type}(M_i) \in \{\text{MLP}, \text{ATTN}\}\}$, Calibration dataset $\mathcal{C}$, Metric $\phi$, Target pruning ratio $\tau$, Number of evaluations in evolutionary search $N$, Search Space $S$. \\
  \ENSURE Pruned model $m^*$
    \STATE $t \gets 0$ \\
    \WHILE{$t < \frac{\tau}{2}$}
        \FORALL{$M_i \in \mathcal{M}$}
            \STATE $S_{i} \gets \text{BlockImportance}(M_i, m, \mathcal{C}, \phi)$
        \ENDFOR
        \STATE $M_{\text{min}} \gets \arg\min_{M_i \in \mathcal{M}} S_{i}$
        \STATE $\mathcal{M} \gets \mathcal{M} \setminus \{M_{min}\}$
        \STATE $t \gets \text{PruningRatio}(m)$
    \ENDWHILE
    \STATE $\text{WeightReordering}(m)$
    \STATE $\{(S_i, \mathcal{M}^{*}_{i}) \}_{i=0}^{N} \gets \text{EvolutionarySearch}(M_i,m,\mathcal{C},\phi,S)$
    \STATE $\mathcal{M} \gets \mathcal{M}^{*}_{\arg\max_{i} \{S_i \mid \text{PruningRatio}(\mathcal{M}^{*}_{i}) == \tau \}_{i=0}^{N}}$
    
    \RETURN $m^*$
  \end{algorithmic}
\end{algorithm}

Algorithm \ref{alg:evol_pruning} describes the steps followed to obtain the pruned model $m^*$. 

\begin{figure}[ht]
    \centering    \includegraphics[width=.46\textwidth]{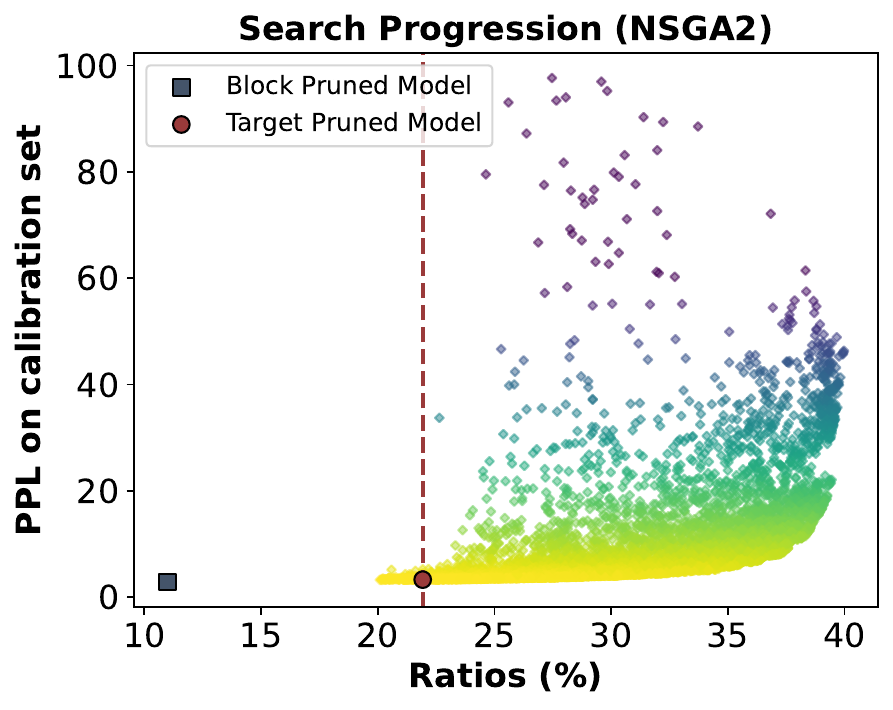}
    \caption{Search progression for evolutionary search (NSGA2) on width dimension (Llama2-7B), starting from a block-pruned model with a pruning ratio of 11\%. The red line is positioned at 22\% (the target pruning ratio). We select the subnetwork with the lowest PPL value on this line as the final pruned model. This search progression figure omits subnetworks with PPL $>$ 100. 
}
\label{fig:search_progression}
\end{figure}

The algorithm begins by pruning residual blocks in the depth dimension until half of the target pruning ratio ($\tau / 2$) is reached. This heuristic provides a balanced starting point for further pruning in the width dimension.
Following the depth pruning, \proj employs the Non-dominated Sorting Genetic Algorithm (NSGA-II) to discover Pareto-optimal pruning configurations for the width dimension. This strategy combines the pruning of MLP channels and attention heads simultaneously, leveraging evolutionary search to achieve a more precise and high-performing pruned model. 
Figure \ref{fig:search_progression} illustrates the search progression for the evolutionary search on the width dimension, starting from a block-pruned model with an 11\% pruning ratio. The plot shows the calibration perplexity (PPL) against the pruning ratio, highlighting the effectiveness of the evolutionary search in identifying optimal configurations.
Specifically, when the target pruning ratio is 22\%, on the dense ratios, as indicated in line 12 of Algorithm \ref{alg:evol_pruning}, the subnetwork with this ratio (within the red line in the figure) and the lowest PPL will be selected as the final pruned model.
As demonstrated in Section \ref{sec:experiments}, this heuristic results in compressed models with similar pruning ratios but better average accuracy than sequential pruning with fixed targets. 
However, it is essential to note that the evolutionary search is computationally expensive and time-consuming to obtain better results.

\subsection{Performance-Recovery Stage} 

To demonstrate the benefits of \proj in production environments, it is essential to consider an additional performance-recovery stage.  
This stage involves fine-tuning the pruned model to recover accuracy on downstream tasks before deployment.

%% file: content/3_experiments.tex
\section{Experiments}
\label{sec:experiments}

\subsection{Setup}

\paragraph{Models} 

To demonstrate the broad applicability of \proj, we conducted experiments using the following models: Llama3.2-3B, Llama3.1-8B, Llama3-8B \cite{dubey2024llama}, Llama2-7B, Llama2-13B \cite{touvron2023llama}, Qwen2.5-7B \cite{yang2024qwen2}, Qwen1.5-7B, Qwen1.5-14B \cite{bai2023qwen}, Baichuan2-7B and Baichuan2-13B \cite{yang2023baichuan}. These models share similar architectures, and their Transformer blocks are composed of self-attention (MHA, GQA \cite{ainslie2023gqa}, etc.) and multilayer perceptrons (MLP). 

\begin{table}[ht]
    \setlength{\tabcolsep}{5pt}
    \centering
    \footnotesize
\renewcommand*\arraystretch{1.1}
    \begin{tabular}{clcc}
    \toprule
        \textbf{Ratio (\%)} & \textbf{Method}  & \textbf{PPL ($\downarrow$)} & \textbf{Avg. Score} \\ 
    \midrule
          \multirow{3}{*}{22} & BlockPruner  &  11.51 & 60.17   \\ 
         ~ & \proj &  \textbf{9.33} & 62.83 \\
         ~ & \proj-Evol &  9.74 & \textbf{63.66} \\
         \cdashline{1-4}
         \multirow{3}{*}{24} & BlockPruner  &  12.16  & 59.01   \\ 
         ~ & \proj &  \textbf{10.01} & 61.99 \\
         ~ & \proj-Evol &  10.61 & \textbf{63.15} \\
         \cdashline{1-4}
         \multirow{3}{*}{27} & BlockPruner  &  14.07 & 55.81   \\ 
         ~ & \proj &  \textbf{11.59} & 60.20 \\
         ~ & \proj-Evol &  12.00 & \textbf{60.79} \\
         \cdashline{1-4}
         \multirow{3}{*}{31} & BlockPruner  &  16.32 & 54.41   \\ 
         ~ & \proj & \textbf{13.77} & \textbf{57.71} \\
         ~ & \proj-Evol &  14.78 & 56.51 \\
    \bottomrule
    \end{tabular}
    \caption{Zero-shot downstream task performance for Llama2-7B with BlockPruner, \proj, and \proj-Evol at different pruning ratios. ``PPL'' refers to the perplexity in Wikitext2. Average score means the average accuracy across five downstream tasks.}
    \label{tab:results_more_pruning_ratios}
\end{table}

\paragraph{Baselines} Building on the analysis by \citet{zhong2024blockprunerfinegrainedpruninglarge}, we include a comparison of \proj with other recent approaches, i.e. SliceGPT \cite{ashkboos2024slicegpt}, LaCo \cite{yang2024lacolargelanguagemodel}, ShortGPT \cite{men2024shortgptlayerslargelanguage} and Relative Magnitude \cite{samragh2023weightsubcloningdirectinitialization_rm}, demonstrating the benefits of the proposed approach.  
In our environment, we reproduced the results of BlockPruner, ensuring the fairness of the competition.

\paragraph{Data and Evaluations} 
For fairness, we follow BlockPruner \cite{zhong2024blockprunerfinegrainedpruninglarge} in using the Alpaca dataset \footnote{https://github.com/tatsu-lab/stanford\_alpaca} as the calibration dataset and employ perplexity as the metric for Block or Width importance. The calibration set consists of either 128 or 256 samples. Regarding evaluation, \proj computes the perplexity (PPL) of Wikitext2 \cite{merity2016pointer}, and then utilizes \emph{lm-eval-harness} \cite{eval-harness} to obtain the zero-shot accuracy on the following downstream tasks: Physical Interaction Question Answering (PIQA)
\cite{Bisk2020_piqa}, Large-scale Winograd Schema Challenge (WinoGrande)
\cite{winogrande}, HellaSwag
\cite{zellers2019hellaswag}, and AI2 Reasoning Challenges (ARC-e, ARC-c)
\cite{Clark2018ThinkYH_arc}. The reader can find all hyperparameters for our experiments in Appendix \ref{sec:hyper_parameters} and in the code repository.

\subsection{Main Results}

The results presented in Tables \ref{tab:main_res} and \ref{tab:main_res2} provide a comprehensive comparison of various pruning methods applied to different large language models, focusing on zero-shot downstream task performance.  The metrics considered include perplexity on Wikitext2 and accuracy scores on five benchmark tasks. The \emph{Dense} row represents the performance of the original, unpruned models.
Our proposed method, \proj, achieves the lowest perplexity and highest average scores across most of the evaluated LLMs, outperforming other training-free pruning approaches. 
For example, regarding the Llama2-7B, \proj achieves the lowest perplexity (9.33) and the highest average score (62.83) among all pruning techniques, indicating superior retention of the model's language understanding capabilities and robustness across diverse tasks. 
Overall, the results demonstrate that \proj offers an effective trade-off between model compression and performance retention, making it a promising solution for deploying large language models in resource-constrained environments.

\subsubsection{Block Pruning in Multiple Dimensions} Notably, \proj outperforms the block-level pruning method, BlockPruner, by employing a more sophisticated approach, block pruning in multiple dimensions. Specifically, \proj subdivides the pruning units into MLP channels and attention heads, allowing for more precise pruning than the coarse-grained block pruning used by BlockPruner or other layer pruning methods. 
The experimental results in the tables demonstrate that this finer granularity leads to more effective model pruning.

\begin{figure}[ht]
    \centering
    \begin{minipage}{0.24\textwidth}
        \centering
        \includegraphics[width=\textwidth]{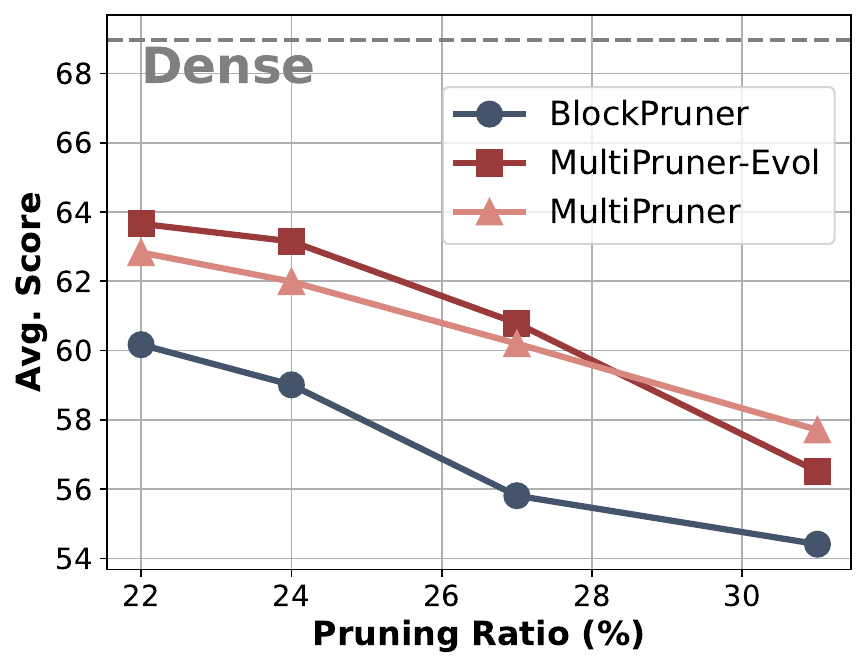}
    \end{minipage}
    \begin{minipage}{0.235\textwidth}
        \centering
        \includegraphics[width=\textwidth]{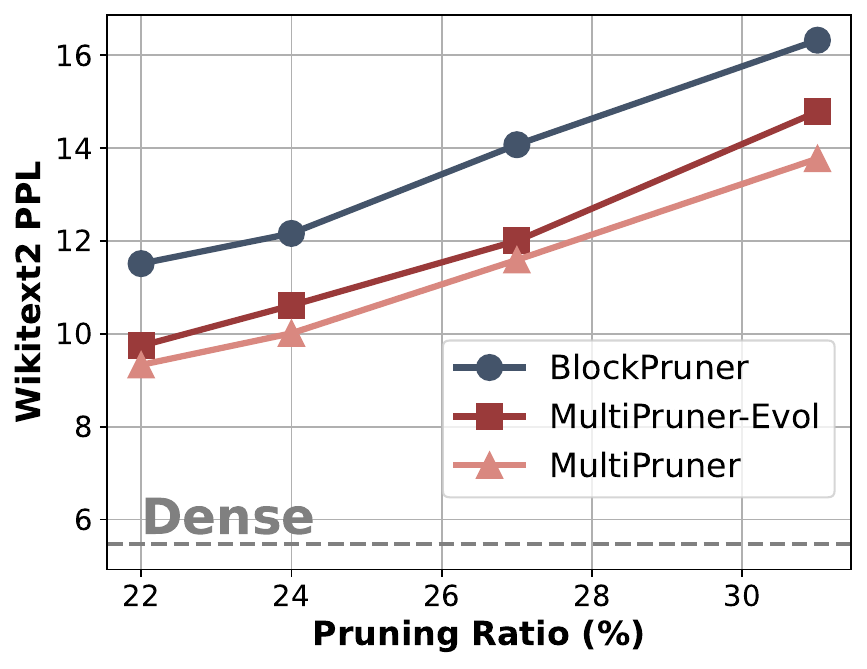}
    \end{minipage}
\caption{Comparison of BlockPruner, \proj and \proj-Evol at different pruning ratios for Llama2-7B. Average score means the average accuracy across five downstream tasks.}\label{fig:more_pruning_ratios}
\end{figure}

\subsection{Exploration of More Pruning Ratios}

As shown in Table \ref{tab:results_more_pruning_ratios} and Figure \ref{fig:more_pruning_ratios}, we explore the performance of the Llama2-7B model using BlockPruner and our proposed method \proj across various pruning ratios (22\%, 24\%, 27\%, and 31\%). 
Across all pruning ratios, \proj consistently achieves lower perplexity and higher average scores than BlockPruner.
For instance, as the pruning ratio reaches 27\%, \proj demonstrates superior performance with a perplexity of 11.59 and an average accuracy score of 60.20. At this pruning ratio, \proj reduces the perplexity of BlockPruner by 2.48 and improves the average score by 4.39\%.
The consistent outperformance of \proj can be attributed to its finer-grained pruning strategy, which allows for more precise and effective pruning, as evidenced by the superior results across various metrics and pruning ratios. The reader can find more results with diverse pruning ratios in Appendix \ref{sec:appendix_more_pruning_retios}.

\subsection{Comparison with Multidimensional Pruning via Evolutionary Search}

In Table \ref{tab:results_more_pruning_ratios} and Figure \ref{fig:more_pruning_ratios}, in addition to comparing BlockPruner and \proj, we also compare the zero-shot downstream task performance of \proj and \proj-Evol.
Both \proj and \proj-Evol outperform BlockPruner, demonstrating the effectiveness of our multidimensional pruning approach. When comparing \proj to \proj-Evol, we observe that \proj-Evol often achieves slightly higher average scores but slightly higher perplexity than \proj.
For instance, at a 24\% pruning ratio, \proj-Evol achieves a higher average score (63.15) compared to \proj (61.99) but a slightly higher perplexity (10.61 vs. 10.01). It is essential to note that \proj-Evol incurs higher computational costs and requires more pruning time. 
In contrast, \proj offers a more balanced trade-off between pruning cost and task performance with lower computational overhead, making it more efficient.

\begin{figure*}[ht]
    \centering
    \begin{minipage}{0.23\textwidth}
        \centering
        \includegraphics[width=\textwidth]{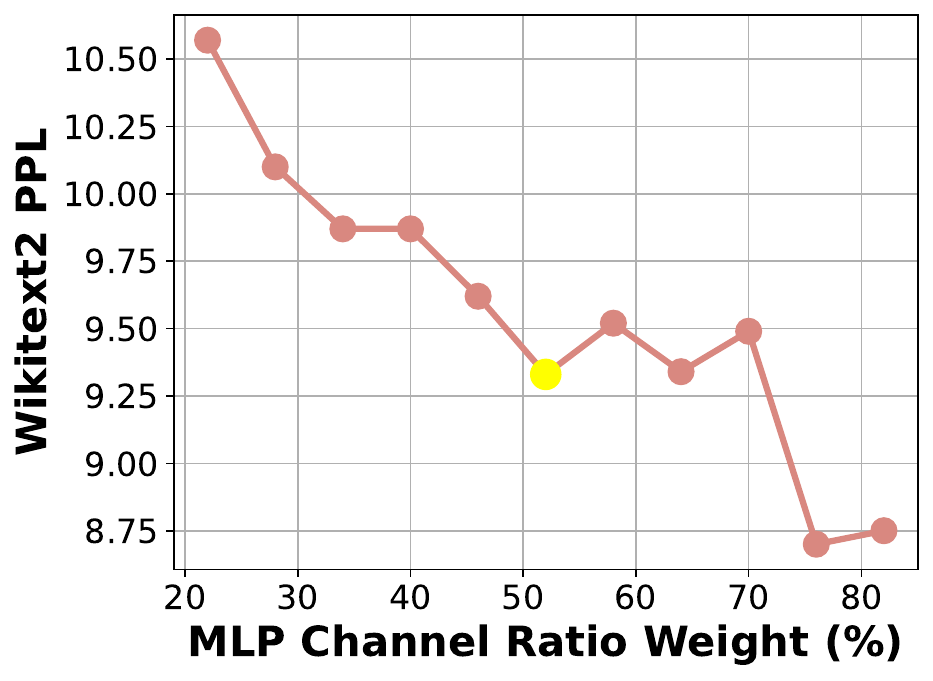}
    \end{minipage}
    \hfill
    \begin{minipage}{0.23\textwidth}
        \centering
        \includegraphics[width=\textwidth]{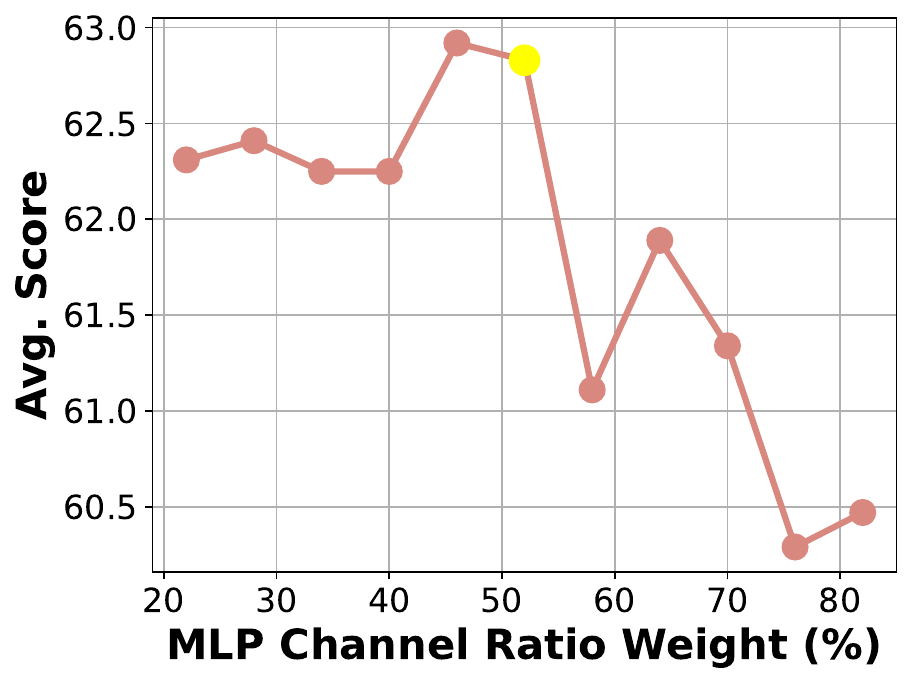}
    \end{minipage}
    \hfill
    \begin{minipage}{0.23\textwidth}
        \centering
        \includegraphics[width=\textwidth]{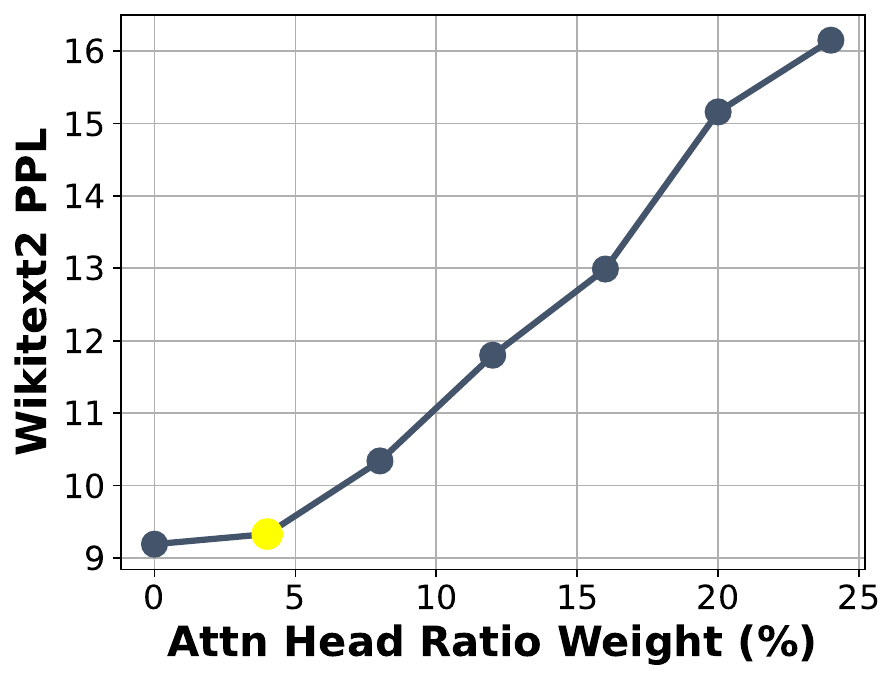}
    \end{minipage}
    \hfill
    \begin{minipage}{0.24\textwidth}
        \centering
        \includegraphics[width=\textwidth]{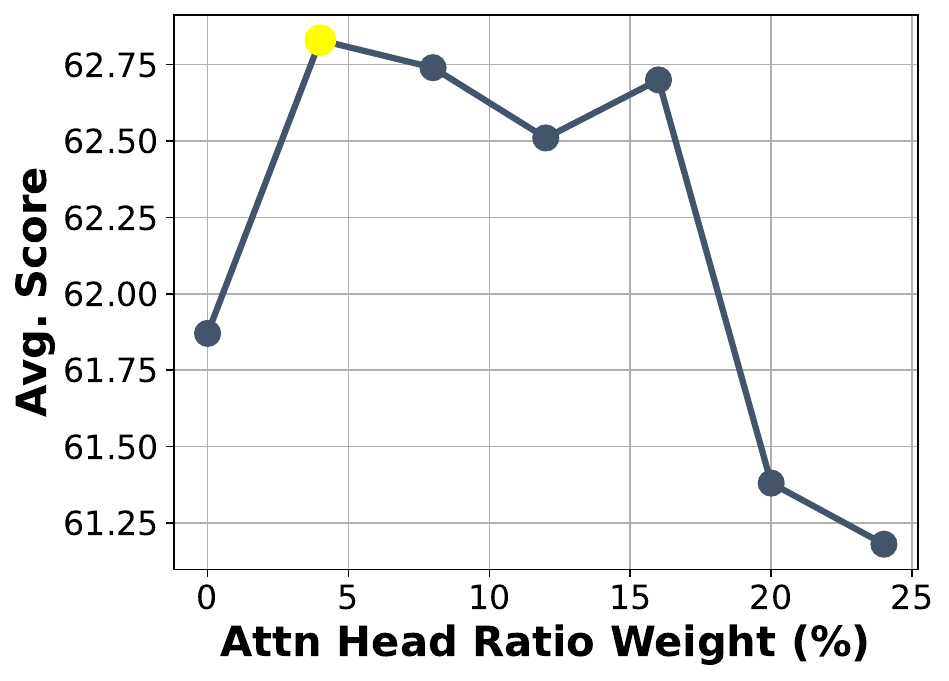}
    \end{minipage}
    \caption{The results of increasing/decreasing the weight of the target pruning ratio allocated to pruning MLP Channels or Attention Heads (Llama2-7B with a target ratio of 22\%). For example, an MLP ratio weight of 50\% means that the pruning ratio for the MLP channel pruning stage is 22\% $\times$ 50\% = 11\%. The yellow point represents the ratio weight we adopted in the most experimental results, which is Block : MLP Channel : Attention Head = 44\% : 52\% : 4\%. Note that the average score means the average accuracy across five downstream tasks.}
    \label{fig:sensitivity}
\end{figure*}

\subsection{Ablation Studies}

\begin{table}[!t]
    \setlength{\tabcolsep}{1.5pt}
    \centering
    \footnotesize 
    \renewcommand*\arraystretch{1.1}
    \begin{tabular}{cccccc}
    \toprule
    \multicolumn{3}{c}{\textbf{Pruning Dimension}} & \textbf{Wikitext2} & \multirow{2}{*}{\textbf{Avg. Score}} \\ 
    \textbf{Block} & \textbf{MLP Channel} & \textbf{Attn Head} & \textbf{PPL ($\downarrow$)} & \\ 
    \midrule
        \cmark(1) & \cmark(2) & \cmark(3) &  9.33 & \textbf{62.83} \\
        \cdashline{1-5}
        \multicolumn{3}{l}{\textit{Order variations:}}\\
        \cmark(1) & \cmark(3) & \cmark(2) &  9.19$_\text{-0.14}$ & 61.87$_\text{-0.96}$ \\
        \cmark(2) & \cmark(1) & \cmark(3) &  9.71$_\text{+0.38}$ & 62.18$_\text{-0.65}$ \\
        \cmark(3) & \cmark(1) & \cmark(2) &  10.26$_\text{+0.93}$ & 61.63$_\text{-1.20}$ \\
        \cmark(2) & \cmark(3) & \cmark(1) &  10.08$_\text{+0.75}$ & 60.14$_\text{-2.69}$ \\
        \cmark(3) & \cmark(2) & \cmark(1) &  9.89$_\text{+0.56}$ & 62.19$_\text{-0.64}$ \\
        \cdashline{1-5}
        \cmark(1) & \cmark(2) & \cmark(2) & 11.25$_\text{+1.92}$ & 62.21$_\text{-0.62}$\\
        \cdashline{1-5}
        \multicolumn{3}{l}{\textit{Existence variations:}}\\
        \cmark(1) & \cmark(2) & \xmark &  9.19$_\text{-0.14}$ & 61.87$_\text{-0.96}$ \\
        \cmark(1) & \xmark & \cmark(2) &  11.04$_\text{+1.71}$ & 61.69$_\text{-1.14}$ \\
        \xmark & \cmark(1) & \cmark(2) &  \textbf{9.02}$_\text{-0.31}$  &  57.79$_\text{-5.04}$ \\
        \cmark(1) & \xmark & \xmark &  11.51$_\text{+2.18}$  &  60.17$_\text{-2.66}$ \\
        \xmark & \cmark(1) & \xmark &   9.41$_\text{+0.08}$ &  57.83$_\text{-5.00}$ \\
        \xmark & \xmark & \cmark(1) &  142.20$_\text{+132.87}$  &  40.60$_\text{-22.23}$ \\
    \bottomrule
    \end{tabular}
\caption{Ablation studies on Llama2-7B + \proj with the pruning ratio of 22\%. Average score means the average accuracy across five downstream tasks. The numbers in the ``Pruning Dimension'' columns indicate the order in which the pruning dimensions are applied. ``\xmark'' indicates that the corresponding pruning dimension is not applied.
}
\label{tab:ablation_studies_stage_order}
\end{table}

In Table \ref{tab:ablation_studies_stage_order}, we present ablation studies on the Llama2-7B model using \proj with a pruning ratio of 22\%. The table examines the impact of pruning dimensions and their pruning order on model performance, considering both perplexity and accuracy scores. 

\textbf{Order Variations:} As shown in the table, the results demonstrate that the default order of Block, MLP Channel, and Attention Head pruning achieves optimal performance.
When the order of the pruning stages is altered, we observe variations in performance. 
For example, starting with MLP Channel or Attention Head generally leads to higher perplexity and lower average scores, confirming that the default order, which moves from coarse to fine granularity progressively, is more effective for precise pruning.

\textbf{Existence Variations:} The existence variations further validate the necessity of each pruning stage. Removing any single stage results in degraded performance. For instance, omitting the MLP channel pruning stage increases perplexity to 11.04 and decreases the average score to 61.69. Removing two stages, such as MLP channel and attention head pruning (i.e., BlockPruner), leads to even more significant performance drops, with perplexity rising to 11.51 and the average score falling to 60.17. Relying solely on attention head pruning results in a drastic increase in perplexity to 142.20 and a significant drop in the average score to 40.60, highlighting that most attention heads are essential and cannot be excessively pruned.

\subsection{Sensitivity Exploration of Self-Attention and MLP}
\emph{What happens when increasing/decreasing the pruning of MLP channels or attention heads (the weight of the overall target pruning ratio)?}
This section explores the sensitivity of MLP channel and attention head pruning.
As shown in Figure \ref{fig:sensitivity}, the four plots illustrate the impact of varying the weight of the target pruning ratio allocated to MLP Channels and Attention Heads on model performance. 

The first two plots show the effect of changing the MLP channel ratio weight while keeping the overall target pruning ratio constant.
As the MLP ratio weight increases from 20\% to 80\%, the Wikitext2 PPL decreases, but the average score initially rises, reaching a maximum of around 50\%, and then declines. This indicates that excessive pruning of MLP channels can negatively impact the model's overall performance.
The latter two plots depict varying weights' influence on the attention head ratio. As this ratio weight increases from 0\% to 25\%, the Wikitext2 PPL consistently rises, particularly when the ratio weight exceeds 5\%.
Similarly, the average score decreases as the head ratio weight increases after 5\%, with a notable drop beyond the 15\% mark. 
This suggests that the attention heads are more sensitive to pruning, and even a slight increase in the pruning ratio can lead to substantial performance loss.

\begin{table}[!t]
    \setlength{\tabcolsep}{6pt}
    \centering
    \footnotesize 
    \begin{tabular}{lccc}
    \toprule
     \textbf{Method} & \textbf{Ratio} & \textbf{PPL ($\downarrow$)} & \textbf{Avg. Score} \\ 
    \midrule
        Dense & 0 & 5.47  & 68.96   \\ 
        \proj & 22\% &  9.33 & 62.83 \\
        \proj \textbf{w/ tune} & 22\% &  \textbf{8.08} & \textbf{64.18} \\
    \bottomrule
    \end{tabular}
\caption{Zero-shot downstream task performance of the compressed Llama2-7B model with recovery tuning (post-training). ``PPL'' refers to perplexity on Wikitext2. The average score is the mean accuracy across five downstream tasks.
}
\label{tab:recovery_finetuning_results}
\end{table}

\begin{table}[!t]
\setlength{\tabcolsep}{6pt}
\footnotesize 
\centering
\renewcommand*\arraystretch{1.2}
\begin{tabular}{lccc}
\toprule
 \multirow{2}{*}{\textbf{Method}} & \multirow{2}{*}{\textbf{Ratio}} & \multicolumn{2}{c}{\textbf{Inference Speedup}}   \\
\cdashline{3-4}
 &  & \textbf{Prefill Stage} & \textbf{Decode Stage}  \\ 
\midrule

Dense & -    & 1.00$\times$ & 1.00$\times$ \\ 
\proj & 22\%    & \textbf{1.32$\times$} & \textbf{1.28$\times$} \\ 
\bottomrule
\end{tabular}
\caption{Inference benchmark results for Llama2-7B (dense vs. pruned). The batch size is 1 and the number of batches is 10. The prompt length is 512. Number of new tokens is 16. The evaluation was conducted on an Intel® Xeon® Platinum 8480+ processor with 56 cores, leveraging Intel® Advanced Matrix Extensions (AMX) to accelerate inference.
}
\label{tab:inference_speedup}
\end{table}

\begin{figure*}[ht]
    \centering    \includegraphics[width=\textwidth]{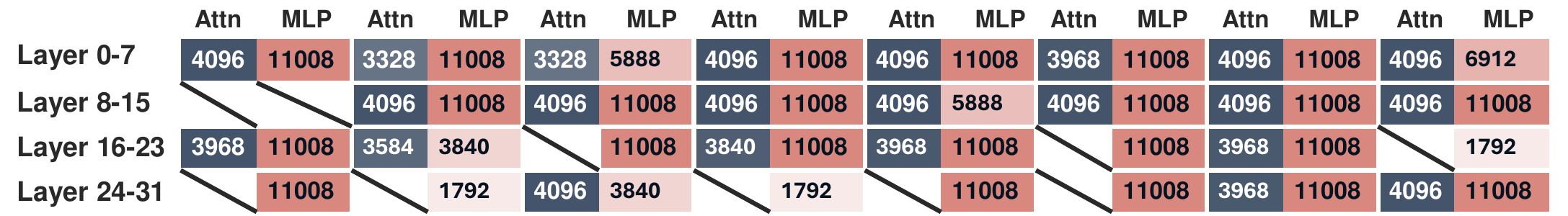}
    \caption{Details of the pruned Llama2-7B model obtained by \proj, including the width of the self-attention and MLP modules across different layers. The numbers within the colored boxes represent the channel sizes, while the white boxes indicate blocks that have been completely removed.}
    \label{fig:details_pruned_llama2}
\end{figure*}
\subsection{Recovery Tuning of the Pruned Model}

Following most of the work \cite{ma2023llmpruner, zhong2024blockprunerfinegrainedpruninglarge}, we also conducted post-training on the pruned model using the cleaned version of Alpaca. The results, shown in Table \ref{tab:recovery_finetuning_results}, indicate significant performance improvements after just two epochs of recovery tuning.
Specifically, \proj \textbf{w/ tune} achieves a reduced perplexity of 8.08 and an increased average score of 64.18.
The recovery tuning phase effectively enhances the performance of the pruned model, making it closer to the original dense model while requiring less computational resources.

\subsection{Inference Speedup}

The inference speedup results for Llama2-7B model, as shown in Table \ref{tab:inference_speedup}, demonstrate the efficacy of the \proj method in enhancing inference performance on an Intel® Xeon® Platinum 8480+ processor. The pruned model, with a 22\% reduction in parameters, consistently outperforms the dense model. 
Specifically, the prefill phase exhibits a speedup of 1.32$\times$, while the decode phase achieves a speedup of approximately 1.28$\times$. 
These results underscore the potential of model pruning techniques to significantly reduce inference latency, facilitating more efficient deployment of large language models in real-world applications. The reader can find additional inference benchmark results across various batch sizes in Appendix \ref{sec:appendix_inference_benchmark_results}.

\subsection{Example of the Pruned Model}

Figure \ref{fig:details_pruned_llama2} illustrates the pruning results for Llama2-7B obtained using \proj. 
We observe that the latter part of the model (layers 16 to 31) is pruned more extensively. 
Notably, the intermediate size of three MLP modules is reduced to 1792, and block pruning is concentrated in the blocks of layers 16 to 31, suggesting that the latter part of the model has more parameter redundancy, while the earlier layers might be more critical. 
Overall, by selectively reducing channel sizes and removing entire blocks where necessary via \proj, the pruned model achieves a more compact architecture without compromising its ability to perform downstream tasks effectively.

%% file: content/4_related.tex
\section{Related Work}
\label{sec:related_w}

The increasing size of large pre-trained models (LPMs) has motivated the development of cost-effective compression techniques to reduce these models' footprint and enable deployment in a broader range of devices. Many methods have been proposed, e.g., pruning \cite{LeCunBrainDamage}, quantization \cite{quantization_survey}, and knowledge distillation \cite{hinton2015distilling}, that overcome the high cost of previous generations of compression algorithms that had fewer resources and time constraints for their execution. Since the paper focuses on pruning, we discuss the evolution of pruning approaches and the latest developments emphasizing efficiency in the compression of LPMs.

\subsection{Pruning Large Pre-trained Models}

Pruning is a popular model compression technique that targets removing or masking redundant elements in a neural network. Pruning methods utilize a \emph{pruning criteria} to accomplish this removal, combined with several strategies to detect the least critical model components efficiently. However, special considerations are required when attempting to prune large pre-trained models. Previous approaches that were successful in pruning small Transformer-based models are not practical for large models, e.g., Movement \cite{sanh2020movement} or Block pruning \cite{lagunas-etal-2021-block}, because they require expensive weights updates. 

\subsubsection{Unstructured Pruning}
Unstructured pruning approaches remove or mask individual weights without any pre-determined pattern. Wanda \cite{sun2023wanda} prunes weights utilizing an unstructured or semistructured strategy by applying a pruning criterion based on the weight's magnitude and the norm of the input activation. BESA \cite{xu2024besa} employs a reconstruction loss per block to sparsify the model. Once sparsity has been induced in the model using any of the mentioned techniques, parameter-efficient fine-tuning techniques (PEFT), e.g., \citet{hu2022lora, munoz-etal-2024-sqft}, can be applied to recover the accuracy for a downstream task. 

Although unstructured pruning can achieve high levels of sparsity, it faces limitations due to the requirement of complex decompression algorithms.

\subsubsection{Structured Pruning}
Structured pruning focuses on removing elements at a higher granularity than unstructured pruning. For instance, in the case of Transformer blocks, these algorithms might remove attention heads or groups of channels in linear layers of the multilayer perceptron (MLP) component.  
The benefits of \emph{structured} pruning are more straightforward to realize than its unstructured counterpart since one can extract the pruned model as a smaller version of the original pre-trained model and utilize the same runtime used by the dense model to realize the benefits and acceleration. 

LoRAPrune \cite{zhang2024loraprune} proposes a structured pruning approach for LPMs that is guided by analyzing the weights and gradients of low-rank adapters (LoRA) \cite{hu2022lora} to determine the importance of components of the Transformer block. 
Recently, several algorithms have been proposed to perform structural removal in a neural network efficiently but without incurring the cost of updating the weights of LPMs. Hence, most state-of-the-art approaches take the \emph{training-free} path. LLMPruner \cite{ma2023llmpruner} removes network structures using gradient information and recovers any accuracy drops utilizing parameter-efficient fine-tuning (PEFT) techniques. 
ShortGPT \cite{men2024shortgptlayerslargelanguage} exploits block redundancy in Transformer-based models and proposes a Block Influence (BI) metric to decide which blocks to prune. BI is a local metric based on the evolution of hidden stages in each block. BlockPruner \cite{zhong2024blockprunerfinegrainedpruninglarge} improves over ShortGPT by proposing a global metric, e.g., the model's perplexity, that is computed by masking the candidate block and assessing its impact if removed. The candidate that results in the most minor drop in performance is removed from the model, which is iteratively conducted until reaching the target pruning ratio. BlockPruner also increases the pruning granularity by focusing on the multi-head attention (MHA) and multilayer perceptrons (MLP), i.e., the minimal residual blocks.

%% file: content/5_conclusion.tex
\section{Conclusion}
Pruning large pre-trained models requires efficient algorithms that consider their immense resource requirements. This paper presents \proj, an efficient training-free structured pruning approach that outperforms other pruning methods. Thanks to their smaller size, the pruned models from \proj accelerate inference and extend the range of devices where machine learning practitioners can deploy these models.

\section*{Limitations}

Due to the complexity of foundation models, the search spaces utilized in our experiments attempt to obtain a good balance between efficiency and efficacy. With a larger computing budget, a finer-grained search might lead to even better results and optimal balancing of the pruning dimensions. Although accuracy is a good indicator of the performance of the pruned models compared to the dense and baseline models, it does not capture the intricacies of these large models. For instance, compressed models are more efficient and have accuracy similar to the base foundation model but might behave differently under certain conditions. Our research focuses on improving the efficiency of large models. Still, additional investigations are required from the larger research community to understand better the impact of the different methods for model compression on the quality of the output from these models.

%% file: content/appendix.tex
\clearpage

\section{Impact of Weight Reordering}
\label{sec:weight_reordering}

\begin{table}
    \setlength{\tabcolsep}{6pt}
    \centering
    \footnotesize 
    \begin{tabular}{lcc}
    \toprule
     \textbf{Weight Importance Metric} & \textbf{PPL ($\downarrow$)} & \textbf{Avg. Score} \\ 
    \midrule
        w/o Reordering &  10.78 & 61.50 \\
        $L^1$-Norm &   \textbf{9.33} & \textbf{62.83} \\
        Wanda \cite{sun2023wanda} &  9.43 & 61.84 \\
    \bottomrule
    \end{tabular}
\caption{Ablation studies for weight reordering on Llama2-7B + \proj with pruning ratio of 22\%. 
}
\label{tab:ablation_studies_weight_reorder}
\end{table}

Table \ref{tab:ablation_studies_weight_reorder} presents the results of ablation studies on the impact of weight reordering before width pruning. The results indicate that applying weight reordering (whether $L^1$-norm or Wanda \cite{sun2023wanda}) significantly improves performance, with the $L^1$-norm performing better than the Wanda strategy. These findings demonstrate that weight reordering enhances the width pruning stage.

\section{Inference Benchmark Results}
\label{sec:appendix_inference_benchmark_results}

Table \ref{tab:inference_speedup_appendix} shows the inference benchmark results for different batch sizes.

\begin{table}
\setlength{\tabcolsep}{5pt}
\footnotesize 
\centering
\renewcommand*\arraystretch{1.1}
\begin{tabular}{clccc}
\toprule
\multirow{2}{*}{\textbf{Batch Size}} & \multirow{2}{*}{\textbf{Model}} & \multirow{2}{*}{\textbf{Ratio}} & \multicolumn{2}{c}{\textbf{Inference Speedup}}   \\
\cdashline{4-5}
 &  &  & \textbf{Prefill} & \textbf{Decode}  \\ 
\midrule

\multirow{2}{*}{1}  & Dense & -    & 1.00$\times$ & 1.00$\times$ \\ 
&\proj & 22\%    & \textbf{1.32$\times$} & \textbf{1.28$\times$} \\ 
\cdashline{1-5}

\multirow{2}{*}{2}  &Dense & -   & 1.00$\times$ & 1.00$\times$ \\ 
&\proj & 22\%    & \textbf{1.46$\times$} &  \textbf{1.29$\times$} \\ 
\cdashline{1-5}

\multirow{2}{*}{4}  &Dense & -    & 1.00$\times$ & 1.00$\times$ \\ 
&\proj & 22\%    & \textbf{1.06$\times$} & \textbf{1.29$\times$} \\ 
\cdashline{1-5}

\multirow{2}{*}{8}  &Dense & -     & 1.00$\times$ & 1.00$\times$ \\ 
&\proj & 22\%    & \textbf{1.15$\times$} & \textbf{1.28$\times$}  \\ 
\cdashline{1-5}

\multirow{2}{*}{16}  &Dense & -    & 1.00$\times$ & 1.00$\times$ \\ 
& \proj & 22\%    & \textbf{1.19$\times$} & \textbf{1.28$\times$}  \\ 
\bottomrule

\end{tabular}
\caption{Inference benchmark results for Llama2-7B (dense vs. pruned). Number of batches is 10. The prompt length is 512. Number of new tokens is 16. The evaluation was conducted on an Intel® Xeon® Platinum 8480+ processor with 56 cores, leveraging Intel® Advanced Matrix Extensions (AMX) to accelerate inference.}
\label{tab:inference_speedup_appendix}
\end{table}

\section{Exploration of More Pruning Ratios}
\label{sec:appendix_more_pruning_retios}
\begin{figure}[ht]
    \centering    \includegraphics[width=.46\textwidth]{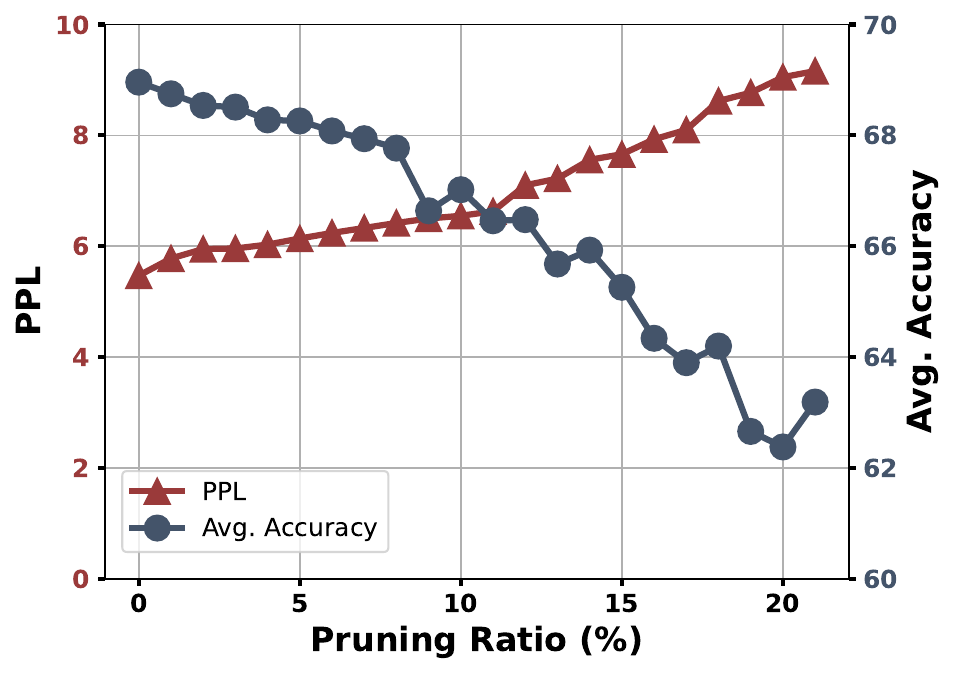}
    \caption{Visualization of Table \ref{tab:pruning_ratio_1_to_21} (performance on zero-shot downstream tasks with pruning ratios ranging from 1\% to 21\%). ``PPL'' refers to the perplexity in Wikitext2. Average accuracy means the average accuracy across five downstream tasks.
}
    \label{fig:pruning_ratio_1_to_21}
\end{figure}

Table \ref{tab:pruning_ratio_1_to_21} and Figure \ref{fig:pruning_ratio_1_to_21} show the results with pruning ratios ranging from 1\% to 21\%.

\begin{table*}[t]
    \centering
    \footnotesize
    \setlength{\tabcolsep}{6pt}
    \renewcommand*\arraystretch{1.3}
    \begin{tabular}{llcccccccc}
    \toprule
        \textbf{Model} & \textbf{Method} & \textbf{Ratio} & \textbf{PPL ($\downarrow$)} &\textbf{PIQA} & \textbf{WinoG} & \textbf{HellaS} & \textbf{ARC-e} & \textbf{ARC-c} & \textbf{Avg. Score} \\ 
    \midrule
        \multirow{22}{*}{\textbf{Llama2-7B}} & Dense & 0\% & 5.47 & 79.05  & 69.06  & 75.99  & 74.54  & 46.16  & 68.96   \\ 
        \cdashline{2-10}
         ~ & \multirow{21}{*}{\proj} & 1\% & 5.78$_\text{+0.31}$ & 78.94 & 68.82 & 75.92 & 74.07 & 45.99 & 68.75$_\text{-0.21}$ \\
         ~ & ~ & 2\% & 5.95$_\text{+0.48}$ & 78.84 & 68.75 & 75.45 & 74.12 & 45.56 & 68.54$_\text{-0.42}$ \\
         ~ & ~ & 3\% & 5.96$_\text{+0.49}$ & 78.51 & 69.22 & 75.33 & 74.20 & 45.31 & 68.51$_\text{-0.45}$ \\
         ~ & ~ & 4\% & 6.03$_\text{+0.56}$ & 78.29 & 68.98 & 75.10 & 73.61 & 45.39 & 68.28$_\text{-0.68}$ \\
         ~ & ~ & 5\% & 6.14$_\text{+0.67}$ & 78.18 & 69.06 & 74.95 & 73.65 & 45.48 & 68.26$_\text{-0.70}$ \\

         ~ & ~ & 6\% & 6.24$_\text{+0.77}$ & 78.02 & 69.30 & 74.57 & 73.19 & 45.31 & 68.08$_\text{-0.88}$ \\
         ~ & ~ & 7\% & 6.33$_\text{+0.86}$ & 77.86 & 69.46 & 74.32 & 73.44 & 44.62 & 67.94$_\text{-1.02}$ \\
         ~ & ~ & 8\% & 6.42$_\text{+0.95}$ & 77.86 & 68.90 & 74.34 & 72.69 & 45.05 & 67.77$_\text{-1.19}$ \\
         ~ & ~ & 9\% & 6.50$_\text{+1.03}$ & 77.64 & 68.59 & 74.27 & 69.36 & 43.34 & 66.64$_\text{-2.32}$ \\
         ~ & ~ & 10\% & 6.55$_\text{+1.08}$ & 77.37 & 68.19 & 74.07 & 71.00 & 44.45 & 67.02$_\text{-1.94}$ \\
         ~ & ~ & 11\% & 6.63$_\text{+1.16}$ & 77.64 & 67.72 & 73.60 & 70.75 & 42.58 & 66.46$_\text{-2.50}$ \\
         ~ & ~ & 12\% & 7.10$_\text{+1.63}$ & 76.33 & 68.43 & 73.77 & 69.82 & 44.03 & 66.48$_\text{-2.48}$ \\
         ~ & ~ & 13\% & 7.22$_\text{+1.75}$ & 76.93 & 68.90 & 73.35 & 67.17 & 42.06 & 65.68$_\text{-3.28}$ \\
         ~ & ~ & 14\% & 7.56$_\text{+2.09}$ & 77.26 & 67.96 & 72.27 & 68.64 & 43.52 & 65.93$_\text{-3.03}$ \\
         ~ & ~ & 15\% & 7.66$_\text{+2.19}$ &76.77&	67.40&	71.82&	68.10&	42.24&	65.26$_\text{-3.70}$ \\
         ~ & ~ & 16\% & 7.93$_\text{+2.46}$ & 76.44 & 66.85 & 71.49 & 66.12 & 40.78 & 64.34$_\text{-4.62}$ \\
         ~ & ~ & 17\% & 8.10$_\text{+2.63}$ & 75.84 & 66.61 & 71.36 & 65.40 & 40.27 & 63.90$_\text{-5.06}$ \\
         ~ & ~ & 18\% & 8.62$_\text{+3.15}$ &75.95&	66.61	&71.45&	65.11	&41.89&	64.20$_\text{-4.76}$ \\
         ~ & ~ & 19\% & 8.77$_\text{+3.30}$ & 75.14 & 64.56 & 69.49 & 63.51 & 40.61 & 62.66$_\text{-6.30}$ \\
         ~ & ~ & 20\% & 9.05$_\text{+3.58}$ & 75.08 & 64.80 & 69.04 & 62.71 & 40.27 & 62.38$_\text{-6.58}$ \\
         ~ & ~ & 21\% & 9.16$_\text{+3.69}$ & 74.81 & 64.80 & 69.35 & 65.53 & 41.47 & 63.19$_\text{-5.77}$ \\
    \bottomrule
    \end{tabular}
\caption{
Performance on zero-shot downstream tasks with pruning ratios ranging from 1\% to 21\%. ``Dense'' denotes the original (unpruned) models. ``PPL'' refers to the perplexity in Wikitext2. 
}   
\label{tab:pruning_ratio_1_to_21}
\end{table*}

\begin{table*}[!t]
    \setlength{\tabcolsep}{6pt}
    \renewcommand*\arraystretch{1.2}
    \centering
    \footnotesize 
    \begin{tabular}{lcc}
    \toprule
\textbf{Hyper-parameter} & \textbf{Value} \\
\midrule
\textit{Pruning Stage:} \\
\cdashline{1-2}
Weight Reorder Importance Metric & $L^1$-norm  \\
MLP Channel Group Size & Hidden size // 4 (e.g., 1024 for Llama2-7B)  \\
Attention Channel Group Size & Head size (one head)  \\
Calibration Dataset & tatsu-lab/alpaca  \\
Block / Channel / Head Importance Metric & Perplexity (PPL)  \\
Number of Calibration Samples for Block Pruning & 256  \\
Number of Calibration Samples for Width Pruning & 128 \\
\midrule
\textit{Recovery Stage:} \\
\cdashline{1-2}
Dataset & yahma/alpaca-cleaned  \\
Epoch & 2 \\
Batch Size & 32 \\
Learning Rate & 1e-4 \\
LoRA Rank & 16 \\
LoRA Alpha & 32 \\
LoRA Dropout & 0.05 \\
LoRA Target Modules & q\_proj, k\_proj, v\_proj, o\_proj, up\_proj, gate\_proj, down\_proj. \\
\bottomrule
    \end{tabular}
\caption{Hyper-parameters used in the experiments. The pruning stage's hyperparameters apply to all LLMs used in our experiment, while the hyperparameters for the recovery stage are specific to Llama2-7B. It is important to note that the pruning ratio weights (distribution) may vary across different LLMs. For instance, in models employing the GQA strategy, \proj generally avoids pruning attention heads due to the smaller number of K and V heads, which are highly sensitive. For more detailed information, please refer to the code repository.
}
\label{tab:hyperparameters}
\end{table*}

\section{Hyper-parameters}
\label{sec:hyper_parameters}
Table \ref{tab:hyperparameters} provides a comprehensive overview of the hyper-parameters used in our experiments, ensuring reproducibility and clarity for the readers. 

For most LLMs in our experiments, the ratio weights we adopted are Block : MLP Channel : Attention Head = 44\% : 52\% : 4\%. For a target pruning ratio of 22\%, this translates to:

\begin{itemize}
    \item Block Pruning Ratio Threshold ($\tau_1$): 22\% $\times$ 44\% = 9.68\%.
    \item MLP Channel Pruning Ratio Threshold ($\tau_2$): 9.68\% + 22\% $\times$ 52\% = 21.12\%.
    \item Attention Head Pruning Ratio Threshold ($\tau_3$): 22.00\%.
    
\end{itemize}